\documentclass[11pt]{article}

\usepackage[preprint]{acl}

\usepackage{times}
\usepackage{latexsym}

\usepackage[T1]{fontenc}

\usepackage[utf8]{inputenc}

\usepackage{microtype}

\usepackage{inconsolata}

\usepackage{graphicx}

\usepackage{booktabs}
\usepackage{makecell}
\usepackage{multirow}
\usepackage{xspace}
\usepackage{paralist}

\newcommand{\framework}{ILA-agent\xspace}       
\newcommand{\benchmark}{Cangjie-bench\xspace}
\newcommand{\execute}{Execute\xspace}
\newcommand{\semsearch}{SemSearch\xspace}
\newcommand{\viewstruct}{ViewStruct\xspace}
\newcommand{\viewdetail}{ViewDetail\xspace}
\newcommand{\typetool}{TypeLookup\xspace}
\newcommand{\submit}{Submit\xspace}

\title{Bridging the Knowledge Void: Inference-time Acquisition of Unfamiliar Programming Languages for Coding Tasks}

\author{
    Chen Shen$^\dagger$ \quad 
    Wei Cheng$^\dagger$ \quad 
    Jingyue Yang$^\dagger$ \quad 
    Huan Zhang$^\dagger$ \quad 
    Yuhan Wu$^\dagger$ \quad 
    Wei Hu$^{\dagger,\,\ddagger,\,}$\thanks{\,\, Corresponding author} \\
    $^\dagger$ State Key Laboratory for Novel Software Technology, Nanjing University, China \\
    $^\ddagger$ National Institute of Healthcare Data Science, Nanjing University, China \\
    \texttt{\{cshen,wchengcs,jyyang,zhanghuan,yhwu\}.nju@gmail.com, whu@nju.edu.cn} 
}

\begin{document}
\maketitle



\begin{abstract}
The proficiency of Large Language Models (LLMs) in coding tasks is often a reflection of their extensive pre-training corpora, which typically collapses when confronted with previously unfamiliar programming languages.
Departing from data-intensive finetuning, we investigate the paradigm of Inference-time Language Acquisition (ILA), where an LLM masters an unfamiliar language through dynamic interaction with limited external resources.
In this paper, we propose \framework, a general ILA framework that equips LLMs with a set of behavioral primitives.
By modeling essential human-like behaviors as a suite of tools, \framework enables LLMs to incrementally explore, apply, and verify language knowledge through structured interactions with the official documentation and execution environment.
To provide a rigorous evaluation in a low-resource setting, we construct \benchmark, a multi-task benchmark based on the novel statically-typed language Cangjie. 
We instantiate \framework for Cangjie and evaluate its performance across code generation, translation, and program repair tasks.
Results using diverse LLMs demonstrate that \framework significantly outperforms retrieval-augmented baselines.
Further analysis of agent trajectories characterizes the emergent behavior patterns while highlighting persisting performance gaps.
\end{abstract}

\section{Introduction}

Large Language Models (LLMs) have demonstrated remarkable proficiency in coding tasks such as code generation \cite{humaneval}, translation \cite{transcoder}, and program repair \cite{Lin2017QuixBugs}. 
This success stems from the internalization of syntactic features and built-in libraries from massive pre-training corpora.
Alongside mainstream Programming Languages (PLs), new general-purpose \cite{cangjie} and domain-specific languages \cite{Mernik2005When} continually emerge to address specialized needs, creating a pressing demand to leverage LLMs for improving developer productivity. 
However, when faced with such unfamiliar PLs, the coding abilities of LLMs degrade sharply \cite{joel2024survey}.

To bridge this gap, a straightforward strategy is to aggregate and augment specialized corpora for finetuning \cite{Wang2025Translating,OptCodeTrans}.
Nevertheless, this paradigm is often hindered by the cold-start challenge inherent in nascent language ecosystems, which requires costly and labor-intensive efforts. 
Alternatively, Retrieval-Augmented Generation (RAG) provides a way to supply LLMs with external context \cite{Lewis2020Retrieval}. 
While it mitigates the rigidity of static parametric knowledge, RAG frequently falters in formulating precise queries for unfamiliar PLs \cite{DocPrompting}.
Furthermore, a purely retrieval-centric approach lacks a mechanism for runtime verification, rendering it inadequate for ensuring the syntactic rigor and complex semantics required for coding tasks \cite{Auto-RAG}.

In this paper, we investigate an active paradigm termed \emph{Inference-time Language Acquisition} (ILA), where a model dynamically learns an unfamiliar PL during the problem-solving process. 
Correspondingly, we propose \framework, a general framework that empowers LLMs as autonomous agents capable of emulating the human-like cognitive process. 
The core of \framework is a suite of behavioral primitives, including (\romannumeral1) Exploration primitives, which enable the model to perform structured semantic queries and navigate official documentation for targeted knowledge acquisition; and (\romannumeral2) Verification primitives, which facilitate interaction with the execution environment to validate the application of language knowledge. 
Furthermore, \framework is designed with tool interfaces for language-specific plugins to augment its primitive capabilities.

To rigorously evaluate ILA capabilities in a low-resource setting, we construct \benchmark, a multi-task benchmark encompassing code generation, translation, and program repair. 
This benchmark is based on Cangjie \cite{cangjie}, a recently released statically-typed language. 
Its novelty and the extreme scarcity of public corpora provide an ideal testbed for assessing ILA. 
We instantiate \framework for Cangjie and integrate a language-specific tool to evaluate its efficacy. 
Experimental results using diverse LLMs \cite{qwen3,deepseekv32,claude} demonstrate that \framework significantly outperforms both task-specific finetuning \cite{OptCodeTrans,Wang2025Translating} and RAG baselines. 
Furthermore, through an in-depth of agent trajectories, we characterize the emergent behavior patterns and pinpoint remaining gaps in current ILA capabilities.

Our main contributions are outlined as follows:
\begin{compactitem}
    \item We propose \framework, a general framework for inference-time language acquisition.
    By integrating exploration and verification primitives, it enables LLMs to incrementally explore, apply, and verify language knowledge.

    \item We construct \benchmark, a rigorous multi-task benchmark encompassing code generation, translation, and program repair.

    \item We conduct extensive experiments to demonstrate that \framework outperforms both task-specific finetuning and RAG baselines, and characterize emergent behavior patterns in agent trajectories.
\end{compactitem}


\section{Problem Formulation}

We first define the available resources of a PL, which can be broadly categorized as follows:
\begin{compactitem}
    \item \textbf{Official documentation:} The authoritative knowledge base providing formal specifications on syntax, characteristics, and built-in libraries, curated by the language designers.

    \item \textbf{Execution environment:} The language's compiler or interpreter, which serves as the ultimate arbiter of code validity by providing definitive feedback on syntactic correctness and runtime behavior.

    \item \textbf{Third-party corpora:} A vast, unstructured collection of existing code from public repositories, forums, and tutorials.
\end{compactitem}

\subsection{Paradigms and Their Challenges}

Approaches to language acquisition can be characterized by the resources utilized and the mechanisms of their deployment.

\textbf{Finetuning}, a data-centric paradigm, primarily relies on the third-party corpora to internalize language knowledge into the model's parameters.
However, the most significant disparity between nascent PLs and established ones lies in the scale and diversity of such corpora. 
Consequently, this paradigm is often hindered by the cold-start challenge inherent in nascent ecosystems, which requires costly and labor-intensive efforts.

\textbf{RAG} provides a training-free way to supply LLMs with external context. 
Its retrieval sources include the official documentation and optional third-party corpora. 
Although this paradigm offers more flexible language acquisition, it faces two primary challenges. 
First, lacking a holistic understanding of an unfamiliar PL, LLMs struggle to formulate effective queries for its unique terms, leading to inefficient and biased retrieval. 
Furthermore, this paradigm lacks a mechanism for interaction with the execution environment, making it overly reliant on the context retrieved prior to a single generation attempt. 
This absence of a verification loop is particularly detrimental for coding tasks in unfamiliar PLs, which demand strict syntactic correctness and complex semantic validation.

\subsection{Inference-time Language Acquisition}

To address these challenges, we investigate ILA, an active paradigm where an LLM masters an unfamiliar language during the problem-solving process. 
In this work, we focus on a foundational setting where the LLM agent is restricted to the minimal viable set of language resources: the official documentation $D$ and the execution environment $E$. 
This choice ensures applicability even when the third-party corpora has not yet established.

We model the ILA process as a partially observable Markov decision process \cite{Kaelbling1998Planning}, where an LLM agent interacts with an environment defined by the language resources $R = \{D, E\}$.
A state $s \in \mathcal{S}$ represents the complete history of the agent's interaction with $R$. 
A state at timestep $t$ consists of the query $Q$ and all past action-observation pairs, denoted by $s_t =(Q,$ $(a_0, o_0), (a_1, o_1), \dots, (a_{t-1}, o_{t-1}))$. 
The state transition is forming $s_{t+1}$ by concatenating the new pair $(a_t, o_t)$ to $s_t$. 
The action space $\mathcal{A}$ comprises a suite of tools that enable the agent to interact with $R$. 
An observation $o \in \mathcal{O}$ is the feedback in response to an action $a \in \mathcal{A}$: $o = R(a)$. 

\begin{figure*}[t]
    \centering
    \includegraphics[width=\textwidth]{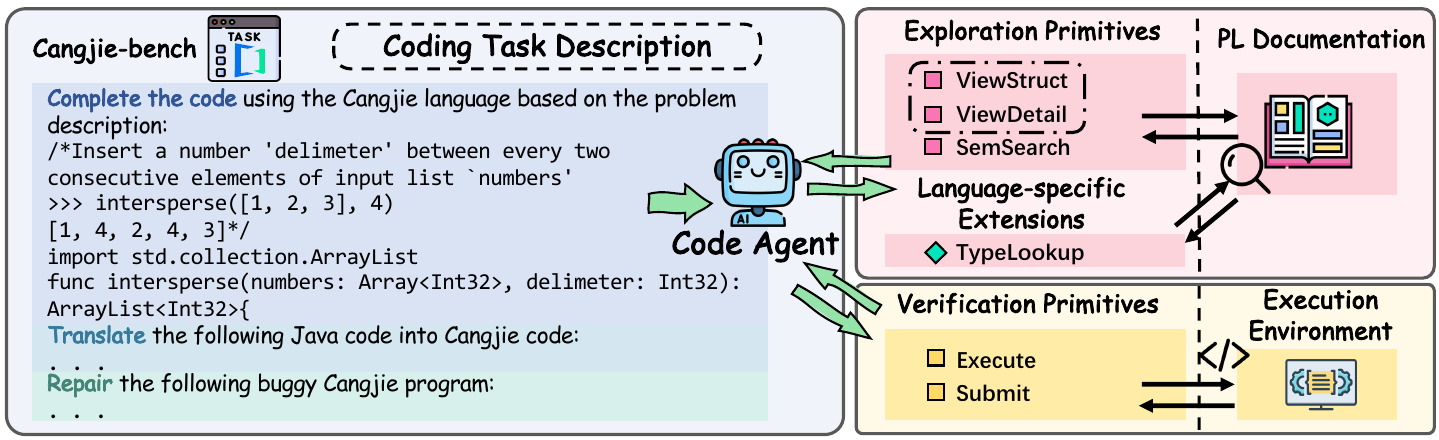}
    \caption{Overview of \framework and \benchmark.}
    \label{fig:framework}
\end{figure*}

The agent's policy $\pi$ is powered by the LLM $\mathcal{M}$. 
At each timestep $t$, the policy maps the current state $s_t$ to an action $a_t \in \mathcal{A}$: $a_t = \pi(s_t; \mathcal{M})$.
A trajectory is the complete instantiation of the ILA process for a given query $Q$. 
It is a finite sequence of states, actions, and observations, starting from the initial state $s_0 = (Q)$ and ending with a terminal action: $\tau = (s_0, a_0, o_0, s_1, \dots, s_T, a_T, o_T)$. 
The trajectory encapsulates the agent's entire path. 


\section{Methodology}

To realize the ILA paradigm, we propose \framework, a general framework that empowers an LLM to master an unfamiliar PL during the problem-solving process. 
Rather than assuming prior familiarity with the target language, \framework is designed to incrementally construct its understanding by exploring documentation, validating language knowledge through execution, and iteratively refining solutions.
As depicted in Figure~\ref{fig:framework}, it equips the agent with a suite of behavioral primitives, implemented as tools, which model the essential human-like behaviors of exploration, application, and verification. 
The entire ILA process operates in a loop, terminating only when the \submit tool reports functional correctness or a maximum number of interaction turns is reached \cite{Zhang2024Pair}. 
The core of \framework consists of exploration primitives, verification primitives, and an interface for language-specific extensions.

\subsection{Exploration Primitives}

The official documentation serves as the primary source of authoritative knowledge. 
It typically includes a user manual detailing the syntax and characteristics, as well as API references for built-in libraries. 
We automatically parse the documentation into structured chunks based on its table of contents and chapter divisions. 
The agent interacts with this structured retrieval source using
\begin{compactitem}
    \item \textbf{Structural views:} We provide two tools that simulate the human behavior of consulting structured documents. 
    They support explicit navigation, allowing the agent to obtain a high-level overview of the documentation's structure with \textbf{\viewstruct}, and dive into a specific section for detailed content with \textbf{\viewdetail}.

    \item \textbf{Global search:} We include \textbf{\semsearch} to perform semantic search across the entire documentation corpus rather than simple keyword matching. 
    This design choice is crucial since an LLM, being unfamiliar with the target language, may generate conceptually-related but lexically-different queries. 
    A robust semantic search is more likely to retrieve relevant information in such cases, whereas keyword matching might return an empty or overwhelmingly large set of irrelevant chunks.
\end{compactitem}

\subsection{Verification Primitives}

The execution environment provides ground-truth feedback, which is essential for validating hypotheses and correcting misunderstandings on language knowledge. 
We equip the agent with
\begin{compactitem}
    \item \textbf{\execute:} This tool accepts and executes any arbitrary code snippet, returning the authentic runtime feedback. 
    During the problem-solving process, it enables the agent to test partial implementations or isolated language constructs. 
    This is critical for verifying unfamiliar language knowledge in a low-stake, exploratory manner.

    \item \textbf{\submit:} This tool accepts a complete solution candidate and evaluates it against a set of predefined public test cases. 
    A successful pass on all public tests serves as a termination condition for the ILA process, indicating that the agent has arrived at a plausible final solution. 
    The ultimate correctness is then determined offline by a separate, private test suite which remains inaccessible to the agent.
\end{compactitem}

\subsection{Language-specific Extensions}

\framework is designed to be extensible, with reserved interfaces for language-specific plugins that can enhance the capabilities of behavioral primitives. 
Such plugins, such as static analysis tools and linters \cite{pylint}, provide more nuanced, language-aware feedback than generic primitives.

To investigate the efficacy of language-specific extensions in our evaluation, we integrate a Cangjie-specific tool called \textbf{\typetool} for exploration. 
Given a type or class name as input, it queries a pre-built index to return the documentation associated with that specific entity, including its members and related descriptions. 
For such a statically and strongly-typed language, this provides a more direct and efficient way for the agent to access type-centric knowledge.



\section{Benchmark Construction}

To rigorously evaluate ILA capabilities in a low-resource setting, we construct \benchmark, a multi-task benchmark encompassing code generation, translation, and program repair. 
The benchmark is founded on Cangjie 0.53.13 \cite{cangjie}, a statically-typed, general-purpose programming language first released on June 21, 2024. 
Its novelty and the extreme scarcity of public corpora ensure that even advanced LLMs \cite{claude} have minimal exposure to its specific syntax and libraries.
As a result, existing LLMs cannot reliably generate syntactically correct code by merely relying on their static, pre-trained knowledge. 
This scenario compels an evaluation of dynamic learning abilities over simple knowledge recall, creating an ideal testbed for assessing the generalized capabilities of an ILA system.
Table~\ref{tab:stats} shows the key statistics of each task in the final suite.
We will release the benchmark publicly.

\subsection{Code Generation}

This task assesses the ability to generate correct Cangjie programs from natural language descriptions.
We develop this subset by manually adapting the HumanEval benchmark \cite{humaneval}.
For each problem, we translate the original Python function signatures, docstrings, and unit tests into their Cangjie equivalents.
We select 155 out of the 164 original problems, excluding 9 tasks reliant on the Python's \texttt{Any} type, since Cangjie's strict static type system does not support such dynamic typing.

\begin{table}
\centering
\resizebox{\linewidth}{!}{
\begin{tabular}{l|ccc}
\toprule
Tasks & Sources & Problems & Avg. Tests \\
\midrule

Code generation     & HumanEval & 155 & \ \ 8.96 \\
Code translation    & TransCoder & 165 & 19.94 \\
Program repair      & QuixBugs & \ \ 32 & \ \ 8.38 \\

\bottomrule
\end{tabular}}
\caption{Statistics of the three tasks in \benchmark.}
\label{tab:stats}
\end{table}

\subsection{Code Translation}

This task measures the ability to migrate source code from a high-resource language to Cangjie.
We adopt the Java-to-Cangjie test set constructed by \citet{Wang2025Translating}, which contains 165 problems derived from the TransCoder dataset \cite{transcoder}. 
To ensure task independence, we deliberately exclude the LowTransEval benchmark \cite{OptCodeTrans}, because its shared origin with HumanEval would introduce overlap with our code generation subset.

\subsection{Program Repair}

This task evaluates the model's capacity to identify and fix bugs in existing Cangjie code.
We construct this dataset by porting the Java programs from the QuixBugs suite \cite{Lin2017QuixBugs}.
For each task, we translate both the buggy Java code and its corrected version to Cangjie.
Unit tests are also translated to ensure they correctly identify the bug and pass on the corrected one.
All original Cangjie code compiles successfully but contains a single functional bug for repair.
We select 32 out of the 40 original problems, excluding 8 tasks due to fundamental differences in language specifications that make a direct translation infeasible.


\begin{table*}
\centering
\small
\begin{tabular}{l|l|cc|cc|cc}
\toprule
\multirow{2}{*}{Models} & \multirow{2}{*}{Approaches} & \multicolumn{2}{c|}{Code generation} & \multicolumn{2}{c|}{Code translation} & \multicolumn{2}{c}{Program repair} \\
\cmidrule{3-8}
 & & ACC & CR & ACC & CR & ACC & CR \\
\midrule

\multirow{2}{*}{Specialized}
 & Cangjie Generator     & 34.84 & 47.10& - & - & - & - \\
 & Cangjie Translator    & - & - & 61.21 & \textbf{85.45} & - & - \\
\midrule

\multirow{4}{*}{DeepSeek-V3.2} 
 & Zero-shot        & \ \ 3.23 & \ \ 5.16 & \ \ 0.00 & \ \ 0.00 & 56.25 & \ \ 71.88 \\
 & Single-time RAG  & 40.00 & 46.45 & 29.09 & 30.91 & 56.25 & \ \ 78.13 \\
 & Iterative RAG    & 47.10 & 52.23 & 41.21 & 50.91 & 68.75 & \ \ \textbf{87.50} \\
 & \framework (ours) & \textbf{63.23} & \textbf{70.97} & \textbf{64.24} & 77.58 & \textbf{71.88} & \ \ \textbf{87.50} \\
\midrule

\multirow{4}{*}{Qwen3-Max} 
 & Zero-shot        & \ \ 7.74 & \ \ 9.03 & \ \ 0.00 & \ \ 0.00 & 62.50 & \ \ 71.88 \\
 & Single-time RAG  & 46.45 & 50.97 & 23.03 & 24.85 & 62.50 & \ \ 78.13 \\
 & Iterative RAG    & 52.90 & 57.42 & 40.61 & 43.64 & 68.75 & \ \ 87.50 \\
 & \framework (ours) & \textbf{73.55} & \textbf{80.65} & \textbf{72.12} & 80.00 & \textbf{81.25} & \textbf{100.00} \\
\midrule

 \multirow{4}{*}{Claude-Sonnet-4.5} 
 & Zero-shot        & 45.16 & 49.68 & 33.94 & 38.18 & 65.63 & \ \ 81.25 \\
 & Single-time RAG  & 59.35 & 63.23 & 36.36 & 40.00 & 62.50 & \ \ 81.25 \\
 & Iterative RAG    & 65.81 & 70.32 & 57.58 & 63.03 & 71.88 & \ \ 84.38 \\ 
 & \framework (ours) & \textbf{81.94} & \textbf{86.45} & \textbf{76.36} & 83.64 & \textbf{90.63} & \textbf{100.00} \\

\bottomrule
\end{tabular}
\caption{Main comparison on \benchmark, where the best results are marked in \textbf{bold}.}
\label{tab:main}

\end{table*}

\section{Experiments and Results}

\subsection{Baselines}

We compare \framework against the paradigms of finetuning and RAG.
See Appendix~\ref{appendix:baseline} for details.

The task-specific finetuning baselines adapt LLMs on Cangjie-specific data, including
\begin{compactitem}
    \item \textbf{Cangjie Generator} is built upon DeepSeek-Coder 6.7B \cite{deepseekcoder}, which contains a Continued Pre-training (CPT) stage on existing Cangjie corpora and a Supervised Fine-Tuning (SFT) stage on our manually curated competitive programming tasks. 

    \item \textbf{Cangjie Translator} \cite{Wang2025Translating} is the state-of-the-art framework specifically for Java-to-Cangjie translation, built upon the Qwen2-7B model \cite{qwen2}.
    Moreover, it leverages the more powerful Qwen2.5-72B-Instruct model \cite{qwen25} to debug and refine the generated code for preference optimization.

    \item \textbf{OptCodeTrans} is a general framework for low-resource language adaptation. 
    Limited by its publicly available resources, we compare against its best published results using StarCoder2-3B \cite{starcoder2}.
    
\end{compactitem}

The training-free baselines include
\begin{compactitem}
    \item \textbf{Zero-shot} serves a lower-bound for the intrinsic language proficiency, which generates code directly from the task description.

    \item \textbf{Single-time RAG} is a standard retrieval-augmented approach where LLMs first formulate a search query based on the problem, then use the retrieved documentation as context for generation \cite{DocPrompting}.
    
    \item \textbf{Iterative RAG} is an advanced, multi-step approach that can iteratively propose new queries to the documentation for refining its understanding \cite{Auto-RAG}.
    We set the maximum number of retrieval rounds to $5$ before producing a final solution.

\end{compactitem}

\subsection{Implementation Details}

We evaluate \framework on three frontier LLMs including DeepSeek-V3.2 \cite{deepseekv32}, Qwen3-Max \cite{qwen3}, and Claude-Sonnet-4.5 \cite{claude}, accessed via their official APIs.
A uniform context window of 128K tokens is used, and the sampling temperature is set to $1.0$.
Each problem-solving trajectory is capped at 15 interaction turns to balance exploration depth and inference cost.
A run terminates early upon a correct submission by the \submit tool.
Otherwise, the final submission is used for evaluation.

Both the \semsearch tool of \framework and RAG baselines share a unified retrieval setup. 
We use Sentence-BERT\footnote{\url{https://huggingface.co/sentence-transformers/all-MiniLM-L6-v2}} \cite{sentence-bert} to generate text embeddings for the documentation chunks. 
For each query content, the retriever returns the top-5 most relevant chunks based on cosine similarity.
All experiments are conducted on Ubuntu 20.04 LTS with a NVIDIA RTX A6000 GPU for local embedding.

\subsection{Evaluation Metrics}

For all coding tasks, we evaluate the accuracy (\textbf{ACC}) of the final solutions. 
A solution is considered correct only if it passes all private test cases.
Furthermore, given that Cangjie is a statically and strongly-typed language, achieving syntactic and type correctness is a fundamental prerequisite. 
We also report the compilation pass rate (\textbf{CR}), which measures the percentage of the final solutions that compile without any errors.

\subsection{Performance Comparison}

The main results are shown in Table~\ref{tab:main}, and an additional comparison on the LowTransEval benchmark is presented in Appendix~\ref{appendix:opt}.
The results demonstrate the consistent and significant superiority of \framework across all tasks and foundation models.

The zero-shot results underscore the effectiveness of \benchmark for evaluating ILA capabilities.
On code generation and translation where the prompt contains no Cangjie code, the performance is near zero across DeepSeek-V3.2 and Qwen3-Max. 
Even the parametric knowledge of the latest LLMs like Claude-Sonnet-4.5 is insufficient for generating valid code from scratch.

By enabling interaction with the official documentation and execution environment, our \framework significantly outperforms both the finetuning and RAG baselines, e.g., an absolute accuracy improvement of 16.13\% to 20.65\% over the iterative RAG on code generation.
This demonstrates that for unfamiliar PLs with extreme data scarcity, activating a model's intrinsic ILA capabilities is more effective than pursuing laborious, task-specific finetuning.
The limitations of the finetuning paradigm are illustrated by the Cangjie translator baseline. 
Despite its sophisticated training pipeline, which involves synthesizing parallel corpora and preference alignment with compiler feedback, it exhibits a severe deficit in functional correctness, even when achieving a high compilation pass rate. 
This highlights the brittleness of such approaches, which often results in superficial pattern matching but fails to generalize to other coding tasks.

Different problem formats also lead to performance variations across three coding tasks.
Code generation and code translation share a fundamental challenge to generating Cangjie code from scratch, i.e., natural language or Java source code, respectively.
On these tasks, retrieval is essential for language acquisition since it is the only source beyond finite parametric knowledge.
In contrast, program repair begins with a nearly complete, syntactically valid Cangjie program, shifting the primary challenge to functional correctness.
Even with a high zero-shot performance due to this rich starting context, the superiority of \framework remains clear.
This demonstrates its reliability as both a code synthesizer and a debugger.

The results demonstrate that the choice of foundation model is a determinant, as evidenced by the varying performance across LLMs.
More capable models are not just better at writing code, but are superior at utilizing the behavioral primitives.
Therefore, \framework also serves as an ILA scaffold for evaluating the agentic capabilities of different LLMs, such as reasoning and tool use.

\subsection{Ablation Study}

\begin{table}[t]
\centering
\small
\begin{tabular}{l|cc}
\toprule
Variants & ACC & CR \\
\midrule
\framework & \textbf{73.55} & \textbf{80.65} \\
\quad w/o Structural views & 65.81 & 70.97 \\
\quad w/o \semsearch & 53.55 & 60.00 \\
\quad w/o \typetool & 57.42 & 61.29 \\
\quad w/o Verification & 53.55 & 59.35 \\
\bottomrule
\end{tabular}
\caption{Ablation results of code generation on \benchmark, using Qwen3-Max.}
\label{tab:ablation-he}
\end{table}

\begin{figure*}[t]
    \centering
    \includegraphics[width=\textwidth]{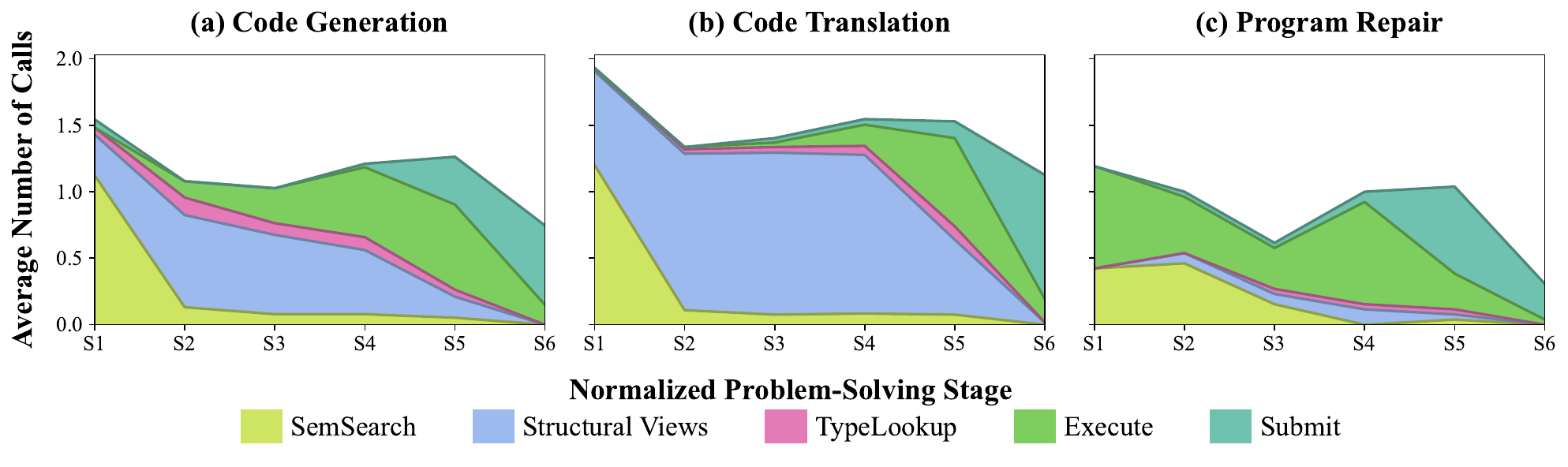}
    \caption{Usage of behavioral primitives across different stages, using Qwen3-Max as the foundation model.}
    \label{fig:stacked_area}
\end{figure*}

To dissect the contribution of each component within \framework, we conduct an ablation study on the code generation task using Qwen3-Max. 
As shown in Table~\ref{tab:ablation-he}, the results systematically quantify the importance of the behavioral primitives.

Removing structural views (i.e., the \viewstruct and \viewdetail tools) leads to the most modest, yet still significant performance drop. 
These two synergistic primitives provide an explicit navigation of the official documentation. 
Although the agent can theoretically use other tools to find specific information, it may struggle to formulate precise and effective queries when faced with unfamiliar PLs.
Structural views mitigate this by allowing the agent to browse the documentation's hierarchical structure, gradually narrowing down to the relevant syntax features or APIs.

The ablations of \semsearch and \typetool highlight the critical need for direct and efficient querying. 
Removing \semsearch, the primary tool for conceptual questions, forces it to rely on slower, multi-step navigation. 
This inefficiency frequently leads to the agent exhausting its maximum interaction budget before arriving at a solution, causing a catastrophic drop in accuracy. 
Similarly, removing \typetool also results in a severe performance penalty. 
As a language-specific tool, it offers a highly structured and reliable way to get precise type information. 
This underscores a key insight for practical applications: equipping ILA agents with language-specific extensions, such as type checkers or fetchers, is a highly effective strategy for boosting both performance and efficiency.

The ablation of the verification primitives (i.e., the \execute and \submit tools) underscores their central role. 
Without the ability to receive feedback from the execute environment, the accuracy falls dramatically to 53.55\%, which is only marginally better than the iterative RAG baseline. 
This minimal performance gain reveals a crucial insight: even with a richer tool set, an agent cannot truly master an unfamiliar PL by only exploring documentation. 
Instead, the true power of exploration tools is unlocked when they operate in concert with the verify-refine loop. 

\subsection{Characterizing Behavioral Primitives}
\label{sec:exp-tools}

To understand how \framework succeeds, we analyze the temporal patterns of tool usage within successful problem-solving trajectories. 
As visualized in Figure~\ref{fig:stacked_area}, we normalize the tool-use sequence of each trajectory into six distinct stages and aggregate the invocation counts of each primitive. 

We first analyze the code generation and translation tasks that require generating Cangjie code from scratch, which show a similar adherence to the three-phase pattern.
It begins with a phase highly dependent on exploration primitives to build foundational knowledge, transitions to a middle phase characterized by a mix of verification and further exploration, and culminates in a final phase focused on verifying the refined code.
As the trajectory progresses, the reliance on \semsearch significantly decreases while structural views rise, which signals a transition from broad exploration to precise reasoning over code structure.
Throughout the entire process, \typetool maintains a low but consistent presence, reinforcing its role as an essential supporting tool for on-demand, high-precision queries, which aligns with our ablation study.

This exploration-first strategy is completely upended in program repair. 
In this task, the primary goal is not exploration but diagnosis, a shift vividly illustrated by the heavy use of the \execute tool from the beginning of the trajectory. 
The agent often first runs the buggy code to obtain the resulting error message, which then inform a more focused exploration phase. 
This demonstrates that the agent intelligently adapts its tool priorities based on the information provided in the prompt, distinguishing between tasks that require broad learning and those that require targeted debugging.

\subsection{Action Transition Dynamics}

\begin{figure*}[t]
    \centering
    \includegraphics[width=\textwidth]{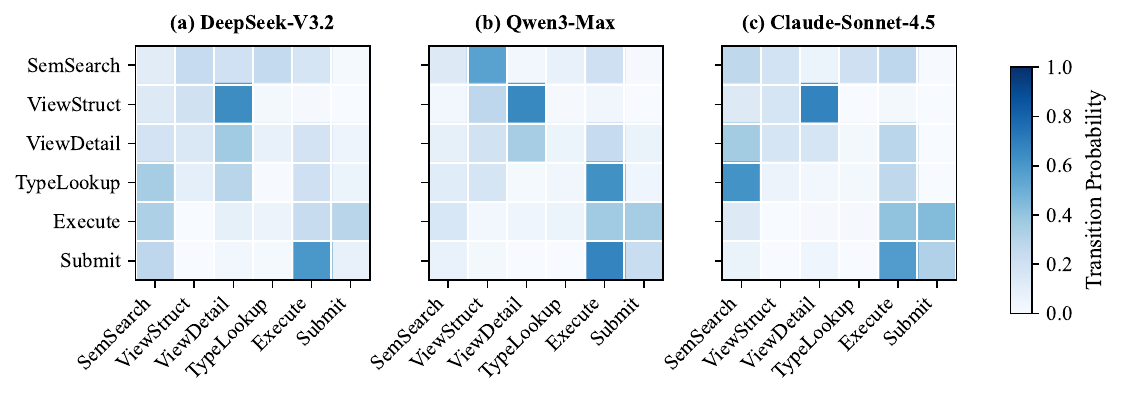}
    \caption{Action transition probabilities across different LLMs, micro-averaged over three coding tasks.}
    \label{fig:tool_transition}
\end{figure*}

To delve deeper into the agent's behavior patterns, we move beyond static tool usage to analyze the dynamics of action transitions.
Figure~\ref{fig:tool_transition} presents the universal strategies of each LLM for coding tasks, visualizing the likelihood of moving from one action to another. 

The analysis reveals that most behavioral patterns are remarkably consistent across LLMs. 
For instance, a strong coupling between structural views is evident, demonstrated by the high transition probability from \viewstruct to \viewdetail. 
Furthermore, the agent exhibits crucial self-verification behavior.
After using exploration primitives, they rarely transition directly to \submit, instead showing a strong tendency to first invoke \execute to validate their code.

However, some differences emerge that reflect unique reasoning styles. 
A notable divergence appears in the action following \semsearch.
Qwen3-Max shows a clear preference for transitioning to \viewstruct, suggesting a strategy of immediately switching from broad search to a more controlled, structured navigation. 
The other two LLMs do not exhibit such preference, indicating a more varied follow-up strategy.
Another key difference is observed after calling \typetool. 
Qwen3-Max tends to immediately proceed to \execute, adopting a direct check-then-use approach to validate the newly acquired type information. 
The other models, particularly Claude-Sonnet-4.5, are more cautious to follow up with another \semsearch, suggesting a safer strategy of seeking further language knowledge before attempting implementation.

Overall, it indicates that all these LLMs exhibit explainable and rational behavioral patterns, even as they have distinct underlying reasoning styles.
Case study of generated Cangjie code and representative agent trajectories is provided in Appendix~\ref{appendix:case_study}.


\section{Related Work}

\subsection{Programming with Unfamiliar PLs}

Adapting LLMs to unfamiliar PLs encompasses two primary research directions: training-based adaptation and inference-time augmentation.
The first direction is mainly the specialized post-training \cite{Efficient,Chen2022On,Cassano2024Knowledge,Giagnorio2025Enhancing,Zhang2025Bridge} and has recently been applied to the Cangjie language \cite{cangjie}. 
OptCodeTrans \cite{OptCodeTrans} utilizes a two-phase tuning process with monolingual CPT and cross-lingual instruction tuning. 
Another study \cite{Wang2025Translating} introduces a training framework using back-translation and preference optimization techniques \cite{Ethayarajh2024kto} to align the model with compiler feedback. 
However, these approaches often rely on the labor-intensive curation of task-specific corpora, which limits their adaptability in other coding tasks \cite{DiscoveryBench}.

To avoid the high costs of model training, the second direction focuses on inference-time methods like RAG approaches \cite{Lewis2020Retrieval,Deligiannis2025RustAssistant}. 
DocPrompting \cite{DocPrompting} establishes a common retrieve-then-generate paradigm by supplying library documentation as context. 
More dynamic approaches have also been explored.
For instance, EVOR \cite{evor} synchronous evolution of queries and a dynamic knowledge base to adapt to unfamiliar syntax during inference. 
Recent studies further explore scaling inference-time computation in unfamiliar domains \cite{Scaling2024,reinforcement,Mora2024Synthetic}. 
Our work contributes to this direction by proposing the ILA paradigm, which focuses on the model's ability to autonomously explore documentation and verify solutions.

\subsection{Agentic Code Intelligence}

Recent progress in code intelligence increasingly adopts multi-turn agentic workflows. 
This shift first emerges in RAG systems to overcome the limitations of passive retrieval \cite{Lu2022ReACC,Cheng2024Dataflow}. 
Active retrieval strategies \cite{Jiang2023Active,DRAGIN} typically trigger retrieval based on token-level uncertainty, while more advanced frameworks \cite{Auto-RAG,Shao2023Iter-RetGen} empower models to autonomously decide when and what to retrieve over multiple turns.

Beyond retrieval, the core idea of agentic frameworks is to close the loop with external feedback \cite{MinionsLLM,zhang2025llms}. 
Foundational frameworks like ReAct \cite{Yao2023ReAct} and Reflexion \cite{Reflexion} demonstrate that interacting with an environment significantly enhances reasoning. 
This paradigm is now widely applied in the coding domain, which effectively handle localized knowledge gaps, such as exploring unseen library APIs \cite{ExploraCoder} or resolving repository-level dependencies \cite{AutoCodeRover,pan2025catcoder,RepoGraph}.
While effective, these studies primarily operate with popular PLs.
In contrast, our \framework is designed for the more fundamental challenge of acquiring an unfamiliar language.


\section{Conclusion}

This paper addresses the critical challenge of extending the coding capabilities of LLMs to unfamiliar PLs. 
We propose \framework, a general framework for inference-time language acquisition that empowers LLMs to master a language by human-like exploration and verification primitives. 
Our experiments on the newly constructed \benchmark demonstrate that \framework significantly outperforms both finetuning and retrieval-augmented approaches. 
Looking ahead, \benchmark provides a rigorous testbed for future efforts in enhancing the model's agentic capabilities, holding significant promise for automating software development in emerging language ecosystems.

\section*{Ethical Considerations}

The datasets and LLMs used in this work are public with permissive licenses.

\section*{Limitations}

While our work demonstrates the effectiveness of the ILA paradigm, we acknowledge several limitations that also frame directions for future research.

First, the performance of \framework is inherently contingent on the quality of the available language resources. 
We assume the existence of well-structured documentation and the execution environment that provide informative feedback, highlighting a symbiotic relationship between AI-driven tools and the language ecosystem itself.

Second, our evaluation is centered on a single, statically-typed language. 
Cangjie is deliberately chosen for its novelty to create a controlled testbed that minimizes the confounding effects of pre-trained knowledge. 
However, this focus implies that the generalizability of our findings to other PLs remains an open question.

Finally, the scope of our \benchmark is intentionally focused on fundamental, self-contained coding tasks. 
This allows us to effectively disentangle the challenge of language acquisition from that of complex algorithmic reasoning. Nevertheless, assessing how these foundational ILA capabilities scale to more intricate, project-level challenges remains a valuable direction for future research.

\bibliography{custom}


\appendix

\section{More Details of Baselines}
\label{appendix:baseline}

This section provides detailed descriptions of the baselines used for comparison in our experiments:
\begin{compactitem}
    \item \textbf{Cangjie Generator}  represents a powerful, data-centric adaptation of the DeepSeek-Coder-6.7B model \cite{deepseekcoder} for Cangjie code generation. 
    First, the model undergoes domain-specific CPT on existing Cangjie corpora, mainly from official and third-party repositories. 
    Then, it is finetuned on a set of 1,200 high-quality, manually curated competitive programming tasks to enhance code generation capabilities. 
    
    \item \textbf{Cangjie Translator} \cite{Wang2025Translating} is the state-of-the-art framework for Java-to-Cangjie translation. 
    It employs a sophisticated offline training paradigm that synthesizes parallel corpora through back-translation and performs preference alignment using compiler-verified feedback to mitigate hallucinations. 
    The framework utilizes the Qwen2-7B model \cite{qwen2} as its foundation for training. 
    For the more demanding steps of data synthesis and LLM-based code refinement, it leverages the more powerful Qwen2.5-72B-Instruct model \cite{qwen25} to debug and refine the generated code, ensuring high syntactic and semantic quality. 
    We directly utilize the performance metrics reported in the original paper for comparison.

    \item \textbf{OptCodeTrans} \cite{OptCodeTrans} is a general framework for adapting LLMs to low-resource programming languages. 
    It consists of a two-phase recipe: (\romannumeral1) CPT on a monolingual corpus of the target language to build foundational understanding, followed by (\romannumeral2) cross-direction instruction finetuning. 
    For our comparison, we refer to the best performance metrics from the original paper, which are produced by training on the StarCoder2-3B model \cite{starcoder2}.

    \item \textbf{Single-time RAG} is a robust RAG baseline following DocPrompting \cite{DocPrompting}. 
    First, in a single query formulation phase, the LLM is prompted with the problem description and instructed to generate up to 5 distinct search queries that it believes are necessary to solve the task. 
    For each query content, we retrieve the top-5 most relevant documentation blocks based on cosine similarity.
    Second, the unique set of all retrieved blocks is aggregated and appended to the original prompt as context. 
    The LLM then performs code generation to produce the final solution. 
    This approach serves as a powerful non-interactive baseline, measuring the model's capacity for comprehensive, single-turn information gathering.

    \item \textbf{Iterative RAG} is an advanced, multi-step RAG variant inspired by the self-reflective mechanisms \cite{Auto-RAG}. 
    It treats the LLM as an active controller that first assesses whether the initially retrieved information is sufficient. 
    If the context is deemed inadequate, the LLM can autonomously propose new, more targeted queries to bridge its knowledge gaps. 
    This iterative refinement process continues until the model feels confident enough to generate the final solution. 
    In our implementation, we allow it to perform up to 5 retrieval rounds, limited by the finite context window for overlong documentation content. 
    For each round, the query mechanism is consistent with the single-time RAG, while the \semsearch tool in ILA-agent can only contain a maximum of three queries at a time.
    
\end{compactitem}

\begin{table}
    \centering
    \small
    \begin{tabular}{l|cc}
    \toprule
    Approaches & ACC & CR \\

    \midrule
    OptCodeTrans  & 45.73 & - \\

    \midrule
    Zero-shot     & \ \ 7.10 & \ \ 8.34 \\
    Single-time RAG     & 48.39 & 53.55 \\
    Iterative RAG     & 54.19 & 61.29 \\
    \framework (ours) & \textbf{74.84} & \textbf{81.29} \\

    \bottomrule
    \end{tabular}
    \caption{Performance comparison with OptCodeTrans on LowTransEval, using Qwen3-Max.}
    \label{tab:lowtranseval}
\end{table}


\section{Comparison with OptCodeTrans}
\label{appendix:opt}
We conduct an additional evaluation on LowTransEval \cite{OptCodeTrans}, which adapts the HumanEval problems \cite{humaneval} into code translation tasks, requiring models to translate correct Java programs into the functionally Cangjie implementation. 
We utilize Qwen3-Max \cite{qwen3} as the foundation model for \framework and RAG baselines, consistent with the analytical experiments presented in the main text. 
We directly report the best accuracy metric of OptCodeTrans provided in the original paper, while they do not provide the compilation pass rate. 
Consequently, we denote this missing value as ``-'' in Table~\ref{tab:lowtranseval}.

The results demonstrate conclusions consistent with the main comparison in Table~\ref{tab:main}.
Additionally, LowTransEval is less challenging than the code translation subset selected for \benchmark.

\begin{figure*}[t]
    \centering
    \includegraphics[width=.95\textwidth]{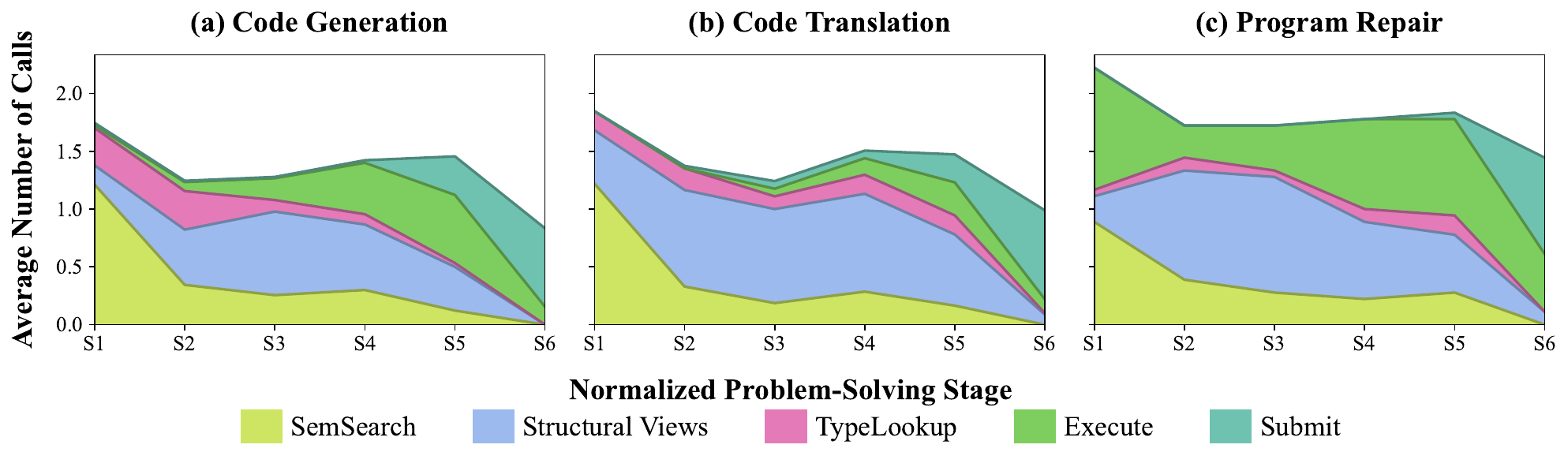}
    \caption{Usage of behavioral primitives across different stages, DeepSeek-V3.2 as the foundation model.}
    \label{fig:stacked_area_deepseek}
\end{figure*}

\begin{figure*}[t]
    \centering
    \includegraphics[width=.95\textwidth]{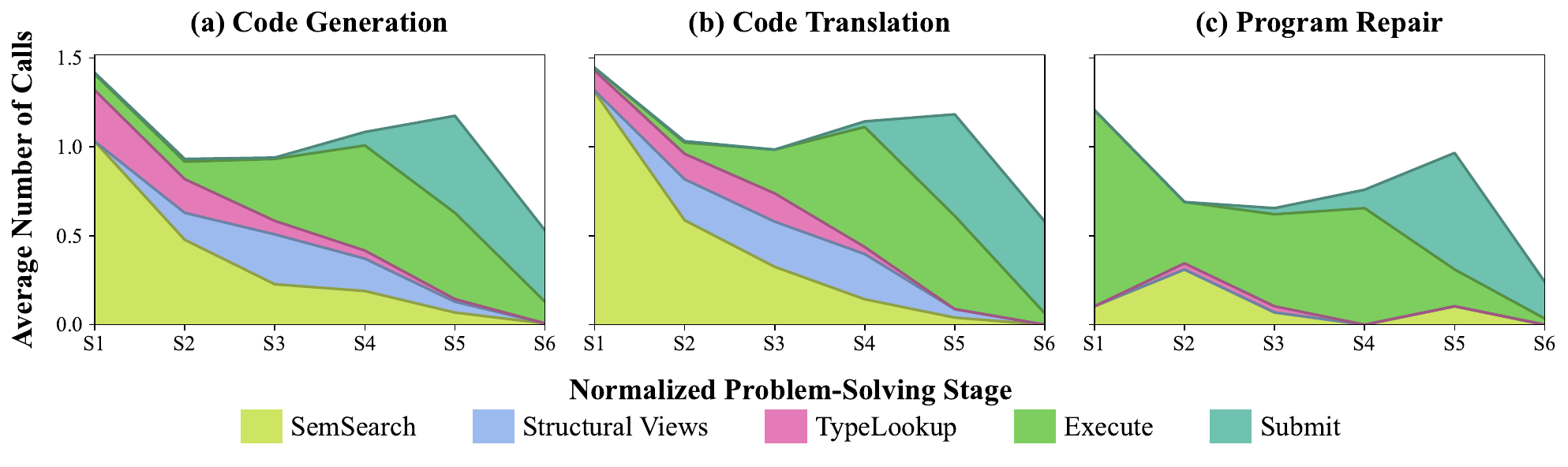}
    \caption{Usage of behavioral primitives across different stages, Claude-Sonnet-4.5 as the foundation model.}
    \label{fig:stacked_area_claude}
\end{figure*}


\section{Behavioral Patterns of Other LLMs}
\label{appendix:more}

To assess the generalizability of the behavioral patterns identified in Section~\ref{sec:exp-tools}, we expand the analysis to include DeepSeek-V3.2 and Claude-Sonnet-4.5. 
As visualized in Figures~\ref{fig:stacked_area_deepseek} and~\ref{fig:stacked_area_claude}, both models consistently follow the broader three-phase acquisition pattern for generation and translation, while also adopting the diagnosis-first approach for repair. 
This alignment confirms that these temporal patterns represent a robust cognitive archetype for ILA-based problem solving.

Despite these shared trends, stylistic variations emerge that reflect the inherent biases of the underlying LLMs.
DeepSeek-V3.2 displays a unique behavior in program repair tasks: it utilizes a high volume of structural views, a pattern absent in the other models. 
This indicates that DeepSeek-V3.2 adopts a ``structurally grounded'' diagnostic strategy, where it attempts to map the entire package or class environment as a prerequisite for localized code fixing. 
Claude-Sonnet-4.5 exhibits a notably lower reliance on structural views, opting instead for a higher density of \semsearch and \typetool invocations. 
This suggests a ``semantic-first'' reasoning style where the model prioritizes conceptual clarity through descriptive documentation over hierarchical exploration.


\section{Case Study}
\label{appendix:case_study}

\subsection{Cangjie Code}

In this section, we present three representative examples of functionally correct Cangjie programs generated by \framework across the three evaluation tasks. 
These implementations highlight the framework's ability to acquire and apply language-specific syntax that is typically absent in generic LLM training data.

\begin{figure}[t]
  \centering
  \includegraphics[width=\linewidth]{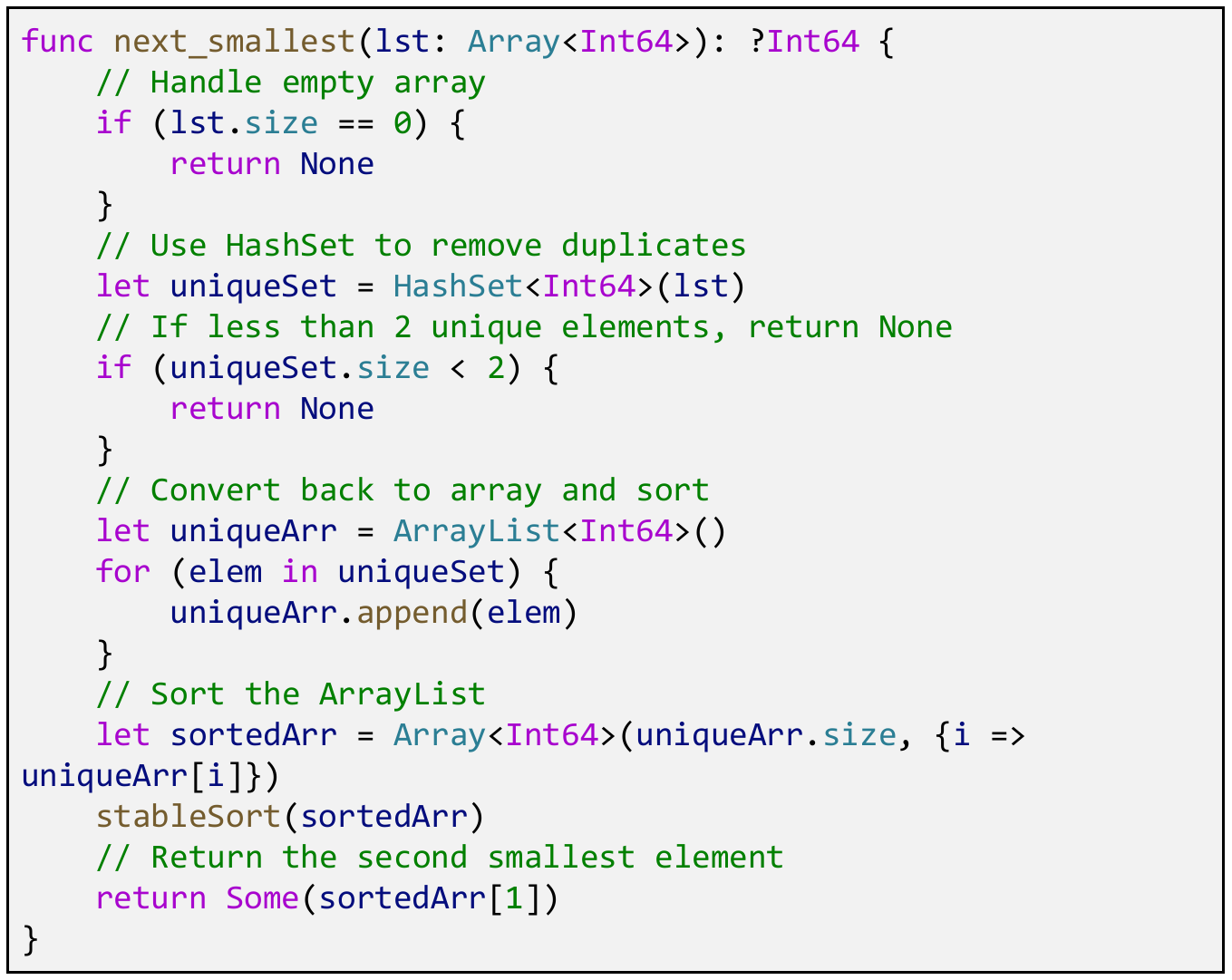}
  \caption{A code generation example.}
  \label{fig:cangjie_case1}
\end{figure}

\paragraph{Code Generation.} As shown in Figure \ref{fig:cangjie_case1}, the model correctly implements the logic to find the second smallest element in an array. 
It demonstrates a mastery of Cangjie's type system by utilizing the \texttt{?Int64} shorthand for the \texttt{Option} type and appropriately returning a value or \texttt{None}. 
Additionally, the code correctly interacts with the \texttt{HashSet} constructor and the \texttt{stableSort} utility from the Cangjie's built-in library to handle duplicate removal and ordering.

\begin{figure}[t]
  \centering
  \includegraphics[width=\linewidth]{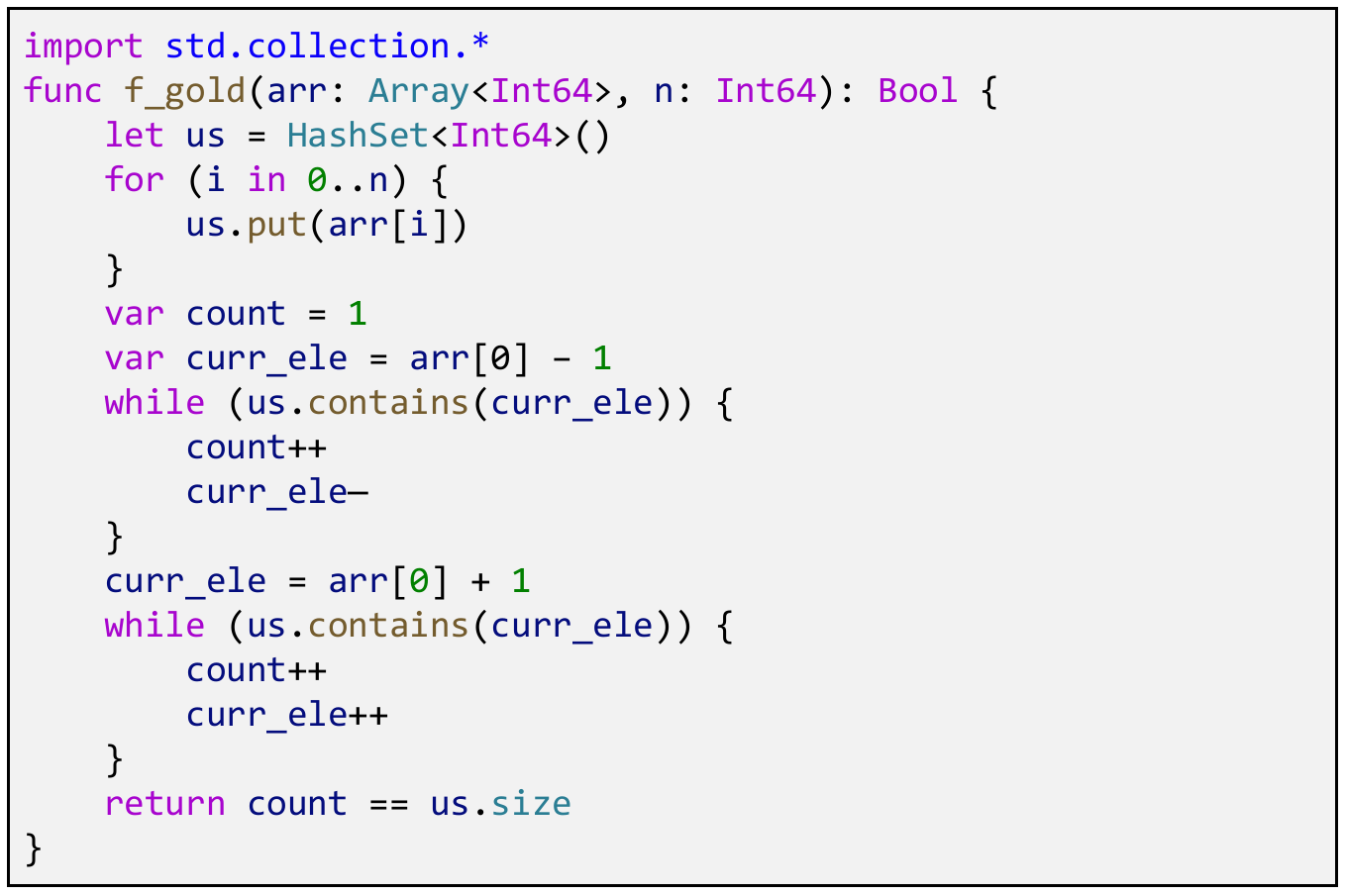}
  \caption{A code translation example.}
  \label{fig:cangjie_case2}
\end{figure}

\paragraph{Code Translation.} Figure \ref{fig:cangjie_case2} shows a translated function that checks if an array's elements form a consecutive sequence. The model successfully adapts Java's collection usage to Cangjie by importing \texttt{std.collection.*} and correctly using the \texttt{put} and \texttt{contains} methods of the \texttt{HashSet} class. It also successfully employs the \texttt{0..n} range expression, which is a signature feature of Cangjie's iteration syntax.

\begin{figure}[t]
  \centering
  \includegraphics[width=\linewidth]{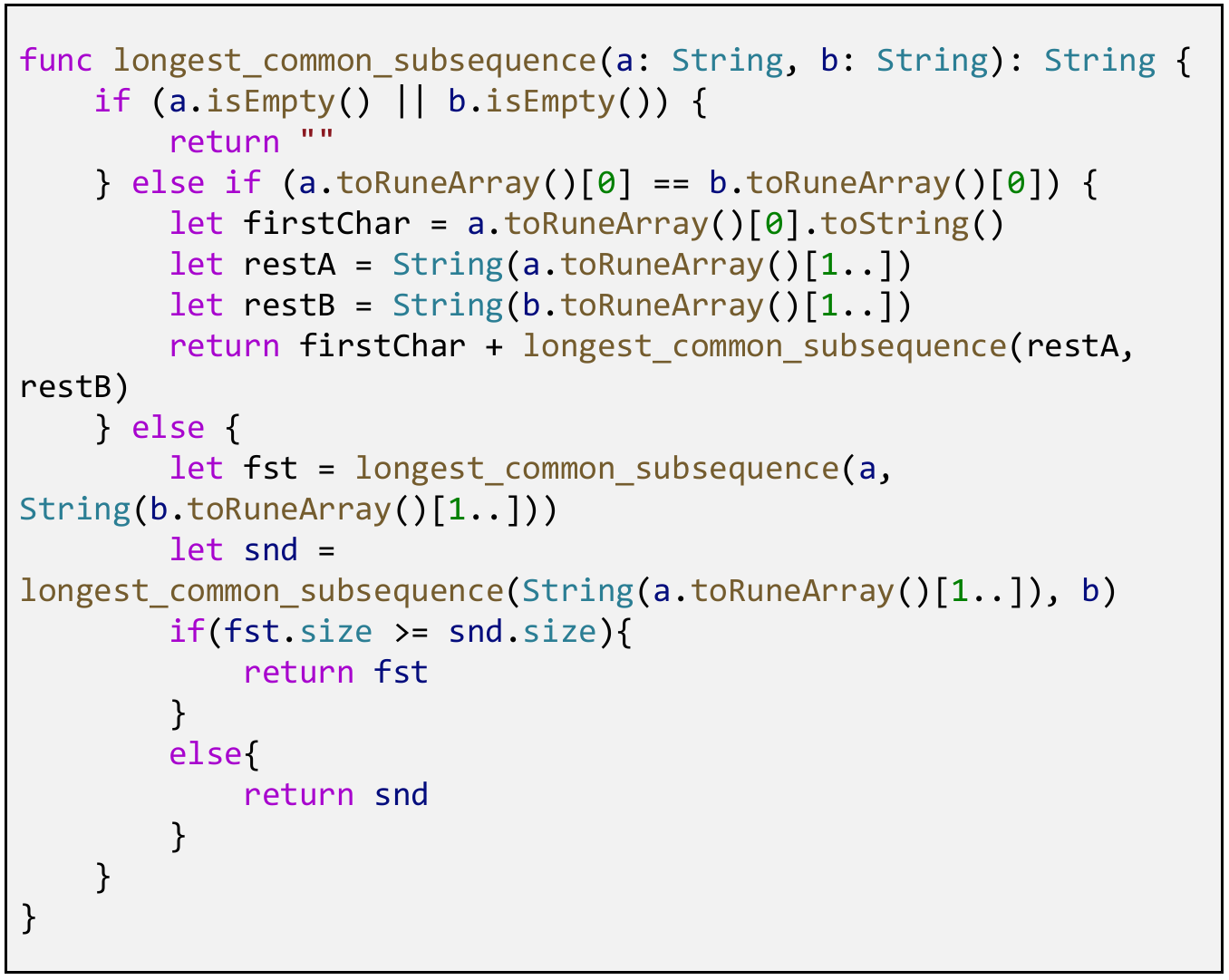}
  \caption{A program repair example.}
  \label{fig:cangjie_case3}
\end{figure}

\paragraph{Program Repair.} In the program repair task (Figure \ref{fig:cangjie_case3}), the model corrects a recursive implementation of the longest common subsequence problem. This example showcases the acquisition of Cangjie's string processing model, where the agent correctly converts \texttt{String} objects to \texttt{toRuneArray()} for character-level indexing and slicing (\texttt{1..}). The precise usage of the \texttt{+} operator for string concatenation and the \texttt{size} property further confirms the model's ability to resolve syntactic discrepancies through execution feedback.

\subsection{Emergent Behavior Analysis}

\begin{figure*}[!t]
    \centering
    \includegraphics[width=.8\textwidth]{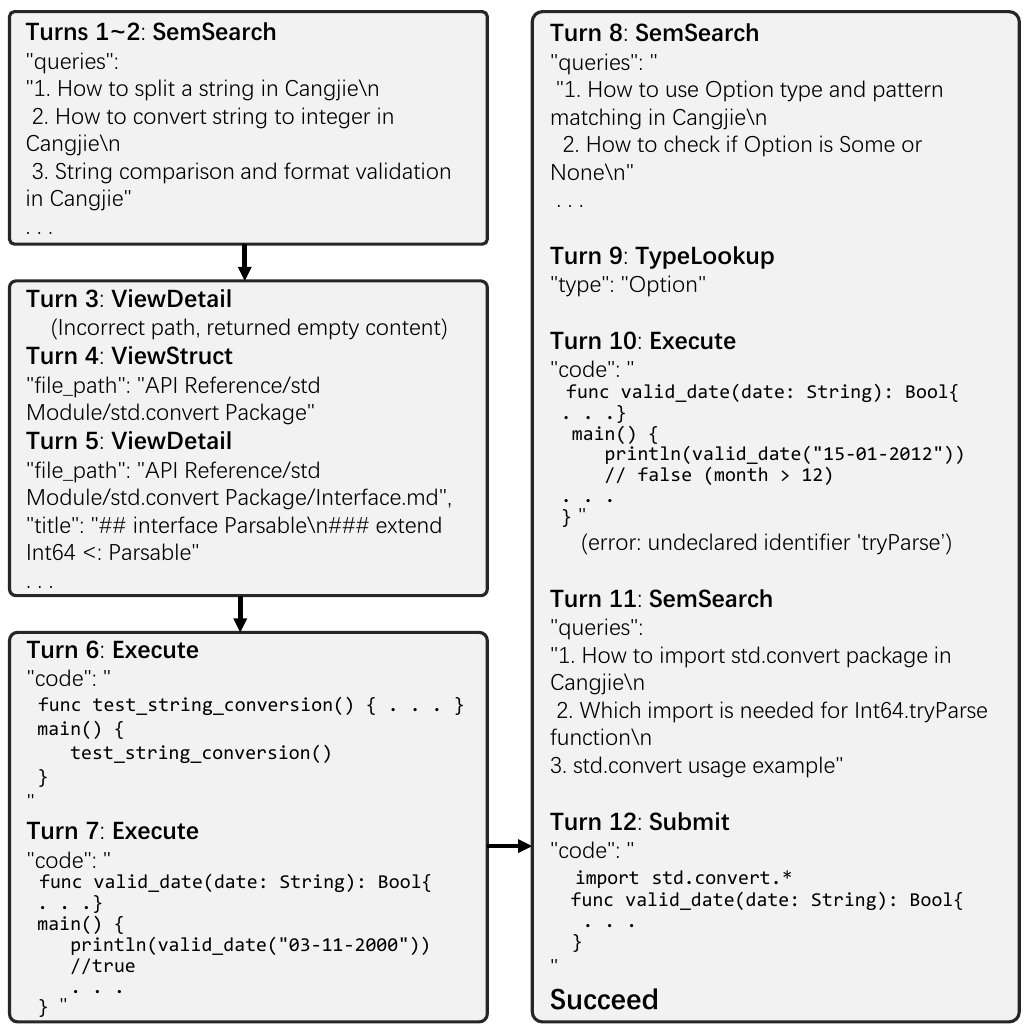}
    \caption{Manifestation of behavioral primitives in a successful ILA trajectory. By interleaving exploration primitives with verification primitives, the agent effectively bridges the knowledge gap of the novel language Cangjie to produce functionally correct code.}
    \label{fig:case_trajectory}
\end{figure*}

To provide a granular view of the ILA process, we present a representative trajectory using Qwen3-Max that successfully solves the HumanEval problem \#124. 
As illustrated in Figure~\ref{fig:case_trajectory}, \framework exhibits a sophisticated interplay between broad semantic acquisition and structured verification through multi-turn interactions.

\paragraph{Phase I: Semantic Knowledge Bootstrapping (Turns 1--2).}
The agent initiates the session with semantic queries via \semsearch. 
It targets fundamental string manipulation and type conversion patterns, e.g., \textit{``How to split a string in Cangjie''} and \textit{``How to convert string to integer in Cangjie''}. This stage aligns the model's general coding intuition with the specific lexical and semantic requirements of the Cangjie's built-in library.

\paragraph{Phase II: Structural Navigation and Recovery (Turns 3--5).}
In Turn 3, the agent attempts to access documentation directly via \viewdetail with an incorrect path, resulting in an empty response. Demonstrating robust error-recovery behavior, the agent subsequently invokes \viewstruct to map the module hierarchy. This structured exploration successfully locates the core conversion interface under \texttt{std.convert/Interface.md}, identifying the \texttt{Parsable} interface and the specific extension \texttt{extend Int64 <: Parsable}.

\paragraph{Phase III: Incremental Functional Verification (Turns 6--7).}
Rather than attempting full implementation immediately, the agent utilizes the \execute primitive in Turn 6 to verify a sub-functional unit: the \texttt{String}-to-\texttt{Int64} conversion logic. In Turn 7, it attempts the complete function but receives execution feedback indicating incorrect handling of the \texttt{Option} type.

\paragraph{Phase IV: Exploration, Refinement, and Final Success (Turns 8--12).}
To resolve the identified gaps, the agent performs a coordinated sequence of \semsearch and \typetool in Turns 8--9. A subsequent execution in Turn 10 fails due to an \textit{``undeclared identifier''} error, revealing a missing \texttt{import} statement. After a final targeted retrieval in Turn~11 to confirm the library path, the agent produces the correct implementation and achieves success via the \submit tool in Turn 12. 

This trajectory exemplifies how \framework enables LLMs to incrementally explore, apply, and verify language knowledge through structured interactions with the official documentation and execution environment.

\end{document}